\documentclass[10pt,twocolumn,letterpaper]{article}

\usepackage{cvpr}
\usepackage{times}
\usepackage{epsfig}
\usepackage{url}
\usepackage[hidelinks,breaklinks=true,bookmarks=false]{hyperref}
\usepackage[utf8]{inputenc}
\usepackage{graphicx}
\graphicspath{{Figures/}}
\usepackage{amsmath}
\usepackage{amssymb}
\usepackage{booktabs}
\usepackage{algorithm}
\usepackage{algorithmic}
\usepackage{subcaption}
\usepackage{float}

\def\confYear{2026}

\title{CACHE-UK: A Stability-Aware Memory Editor for
       Sequentially Updated Quantized LLMs in Finance}

\author{
Anubhav Lakra\textsuperscript{1}\quad
Yue Feng\textsuperscript{2}\\
\textsuperscript{1}Indian Institute of Technology Madras, Chennai, India\\
\textsuperscript{2}University of Birmingham, Birmingham, United Kingdom\\
{\tt\small a4anubhavlakra@gmail.com, y.feng.6@bham.ac.uk}
}

\begin{document}
\maketitle

% -  -  -  -  -  -  -  -  -  -  -  -  -  -  -  -  -  -  -  -  -  -  - -
\begin{abstract}
Large Language Models (LLMs) deployed in dynamic financial environments
face a critical challenge: maintaining factual accuracy as market
conditions, regulations, and corporate facts change continuously.
While 4-bit quantization enables efficient deployment, it severely
limits the viability of sequential memory editing-existing methods
undergo catastrophic performance degradation under this ``quantization
stability crisis.'' We introduce CACHE-UK (Contextual Adaptive
Continual Hybrid Editor for UK Finance), a stability-aware memory
editing framework specifically designed for domain-specific, quantized
LLMs. CACHE-UK integrates three components: a rank-1 LoRA perturbation
mechanism that confines edits to the low-rank adapter subspace, a
financial domain prioritization module for content-adaptive edit
strength, and a closed-loop Stability Controller that tracks
``degradation debt'' to prevent catastrophic forgetting across
sequential updates. Evaluated on a 4-bit quantized OpenLLaMA-3B model
with a curated UK financial corpus of 88{,}021 documents, CACHE-UK
reduces knowledge degradation by 11--17\% relative to adapted
baselines under identical 4-bit constraints - its most robust
effect - while attaining the highest test success (generalization)
rate observed in our setting (28\%, a 6 percentage point improvement
over the strongest adapted baseline). These results indicate that stability-aware editing can
improve factual maintenance in resource-constrained financial LLM
deployments, though absolute generalization rates remain low.
\end{abstract}

% -  -  -  -  -  -  -  -  -  -  -  -  -  -  -  -  -  -  -  -  -  -  - -
\section{Introduction}

\begin{figure*}[t]
  \centering
  \includegraphics[width=0.92\textwidth]{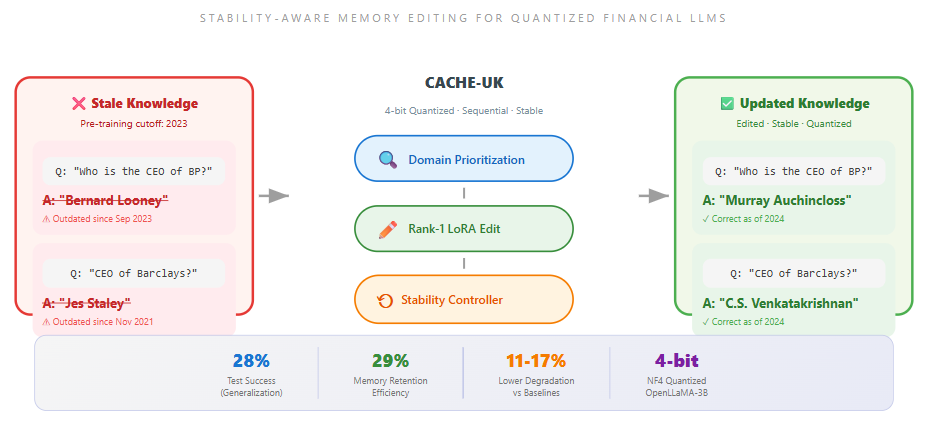}
  \caption{CACHE-UK overview. Stale financial knowledge (left) is updated via domain prioritization, rank-1 LoRA perturbation, and degradation debt control under 4-bit quantization (right).}
  \label{fig:teaser}
\end{figure*}

The deployment of Large Language Models (LLMs) in financial services
has grown rapidly, driven by their capacity for complex reasoning, fact
retrieval, and language understanding~\cite{brown2020language,vaswani2017attention}.
Yet these models embed a static snapshot of world knowledge from
training time. In the UK financial sector - where regulatory guidance,
executive appointments, interest rate policies, and company filings
change on a daily basis - this staleness poses an operational
risk~\cite{wang2024memory,ke2025demystifying}.

Consider a concrete example: when BP's CEO changed from Bernard Looney
to Murray Auchincloss in September 2023, any LLM trained before that
date would confidently output the wrong answer to ``Who is the CEO of
BP?'' - a factual error with real consequences for automated financial
analysis, compliance reporting, and client-facing advisory systems.
Such changes occur across hundreds of UK-listed companies each year,
making periodic retraining impractical.

Full-model fine-tuning can incorporate new data, but it is
computationally prohibitive, risks catastrophic forgetting of general
capabilities~\cite{mccloskey1989catastrophic}, and cannot realistically
keep pace with the cadence of financial updates. Memory editing has
emerged as a computationally efficient alternative: rather than
retraining, it surgically modifies specific facts stored within a
model's parameters~\cite{yao2024comprehensive,wang2023easyedit}.
Methods such as ROME~\cite{meng2022locating}, MEMIT~\cite{meng2023mass},
and MEND~\cite{mitchell2022fast} have demonstrated success on
full-precision models. However, practical deployment increasingly relies
on 4-bit quantization for resource efficiency~\cite{dettmers2023qlora,frantar2022gptq}, and the interaction between memory editing and
quantization remains critically underexplored.

We identify and empirically characterize the \textit{quantization
stability crisis}: a phenomenon in which sequential edits on
low-precision models cause cascading memory degradation. Quantization
maps continuous weights to a discrete grid, creating a
\textit{resolution barrier} - a surgical rank-one update may be smaller
than the quantization step size, causing it to be rounded to zero. This
drastically narrows the ``stability margin'' for any edit, and repeated
edits compound the interference.

To address this, we propose \textbf{CACHE-UK} (Contextual Adaptive
Continual Hybrid Editor for UK Finance), a modular framework for stable,
sequential knowledge updates in quantized, domain-adapted LLMs. Our
contributions are:
\begin{itemize}
  \item A characterization of the quantization stability crisis both
        analytically - the 4-bit quantization step size is 16$\times$
        coarser than 8-bit - and empirically: all adapted baselines
        collapse to 12--22\% test success under 4-bit quantization.
  \item A three-component framework integrating rank-1 LoRA
        perturbation (operating on rank-16 adapter matrices with 51K
        parameters rather than full weight matrices with 10.2M),
        financial domain prioritization, and a closed-loop Stability
        Controller governed by a ``degradation debt'' metric.
  \item Empirical evaluation on a curated UK financial corpus of
        88{,}021 documents, demonstrating improved performance over
        the adapted baselines we evaluate in test success (+6pp) and
        degradation reduction (11--17\%).
  \item A documented pipeline from corpus curation through domain
        adaptation to stability-aware editing, applicable to
        LoRA-adapted quantized models in regulated domains.
\end{itemize}

% -  -  -  -  -  -  -  -  -  -  -  -  -  -  -  -  -  -  -  -  -  -  - -
\section{Background and Related Work}

\subsection{Memory Editing Methods}

Memory editing seeks to modify factual knowledge in LLMs without full
retraining, grounded in the observation that factual associations are
localized in feed-forward networks (FFNs) acting as key-value
memories~\cite{geva2020transformer}.

\textit{Solver-based methods} compute a direct weight update. ROME
\cite{meng2022locating} applies a rank-one update to a causal-traced
critical layer; MEMIT~\cite{meng2023mass} generalizes this to batch
edits via a constrained least-squares formulation across multiple layers.

\textit{Meta-learning frameworks} train an auxiliary editor network.
MEND~\cite{mitchell2022fast} learns to predict low-rank updates from
factual error gradients. KnowledgeEditor~\cite{decao2021editing}
uses a hyper-network with constrained optimization to modify facts
while minimizing collateral damage.

Other approaches include representation engineering for steering model
behaviour through activation manipulation~\cite{zou2023representation}
and toolkit frameworks such as EasyEdit~\cite{wang2023easyedit} that
unify multiple editing strategies under a common API. All of these
methods target full-precision models and have not been evaluated under
post-training quantization - the setting we address.

\subsection{Quantization and Efficient Deployment}

Post-training quantization (PTQ) techniques such as
GPTQ~\cite{frantar2022gptq} and QLoRA~\cite{dettmers2023qlora} have
made 4-bit inference practical for large-scale deployments. Low-Rank
Adaptation (LoRA)~\cite{hu2021lora}, which injects trainable low-rank
matrices while freezing base weights, has become the standard PEFT
approach for domain adaptation~\cite{houlsby2019parameter}. Our
framework operates natively within the LoRA adapter subspace, making
it directly compatible with quantized deployments without requiring
dequantization of the frozen base weights.

\subsection{Continual Learning and Catastrophic Forgetting}

Sequential updates in neural models risk catastrophic forgetting - the
overwriting of previously learned knowledge~\cite{mccloskey1989catastrophic}.
Continual learning methods typically address this through
regularization (e.g., EWC) or rehearsal; however, these are not
designed for the specific setting of surgical, fact-level edits under
quantization constraints. CACHE-UK addresses this gap via a closed-loop
controller inspired by integral control theory.

% -  -  -  -  -  -  -  -  -  -  -  -  -  -  -  -  -  -  -  -  -  -  - -
\section{The Quantization Stability Crisis}

\subsection{The Resolution Barrier}

Quantization maps the continuous weight space to a discrete set of
values, creating a \textit{resolution barrier}. For $b$-bit
quantization with range $R$, the step size is
$q = R / (2^b - 1)$. A rank-one update $\Delta W$ with
$\|\Delta W\|_\infty < q$ is effectively rounded to zero, nullifying
the edit entirely. For 4-bit quantization ($b=4$, $q = R/15$), this
step size is \textbf{16$\times$ larger} than for 8-bit ($b=8$,
$q = R/255$), drastically narrowing the window of edit strengths that
are both large enough to register and small enough not to cause
catastrophic interference.

\begin{figure}[t]
  \centering
  \includegraphics[width=\columnwidth]{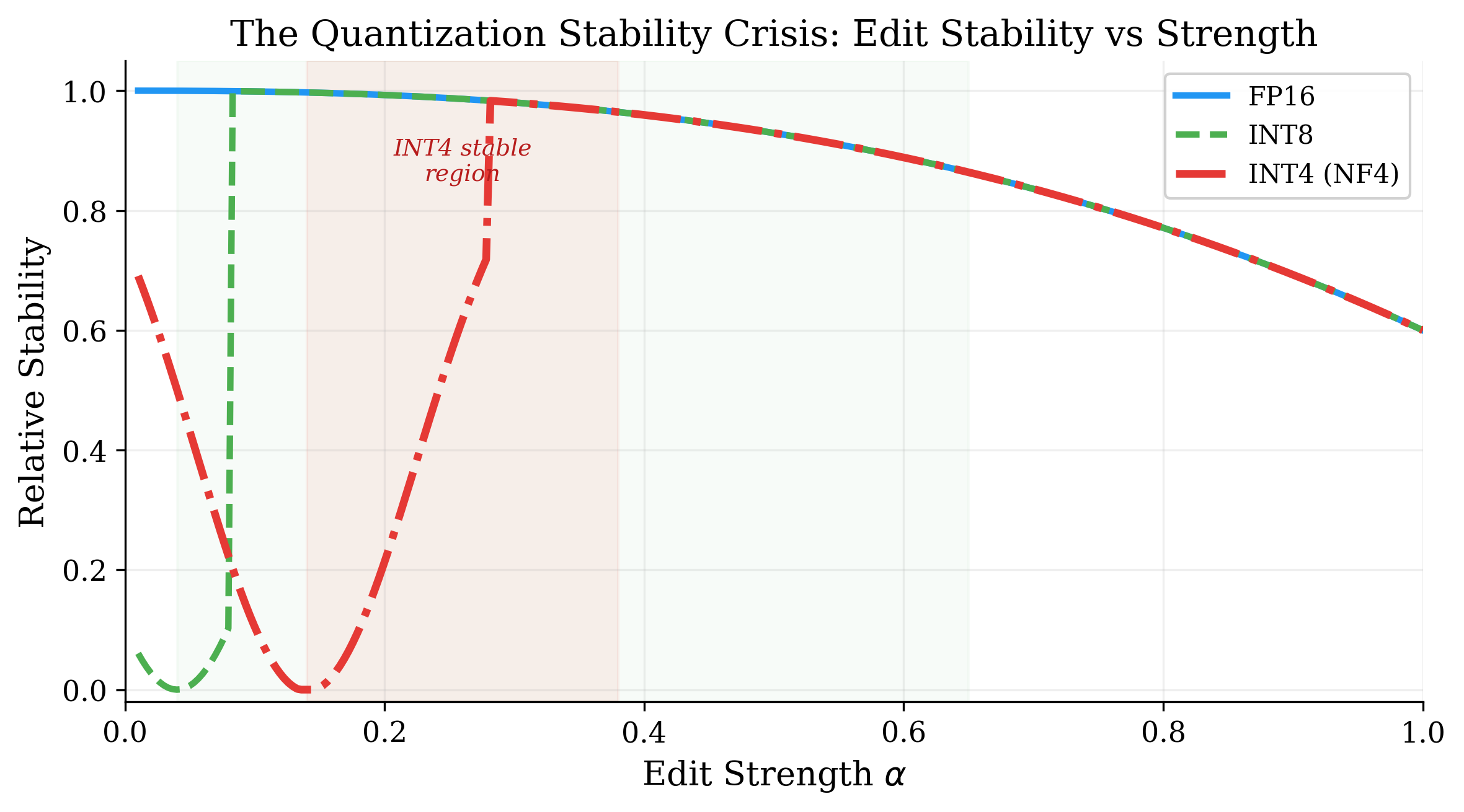}
  \caption{Conceptual schematic (not measured data): an illustrative
           model of edit stability across precisions. INT4 quantization
           narrows the stable edit-strength region relative to INT8 and
           FP16. The shaded region indicates the approximate viable
           operating range for 4-bit editing.}
  \label{fig:stability_margin}
\end{figure}

Figure~\ref{fig:stability_margin} illustrates this phenomenon: the
stable region of edit strengths is dramatically narrower for INT4 than
for higher-precision formats.

\subsection{Inter-Edit Interference}

\begin{figure}[t]
  \centering
  \includegraphics[width=\columnwidth]{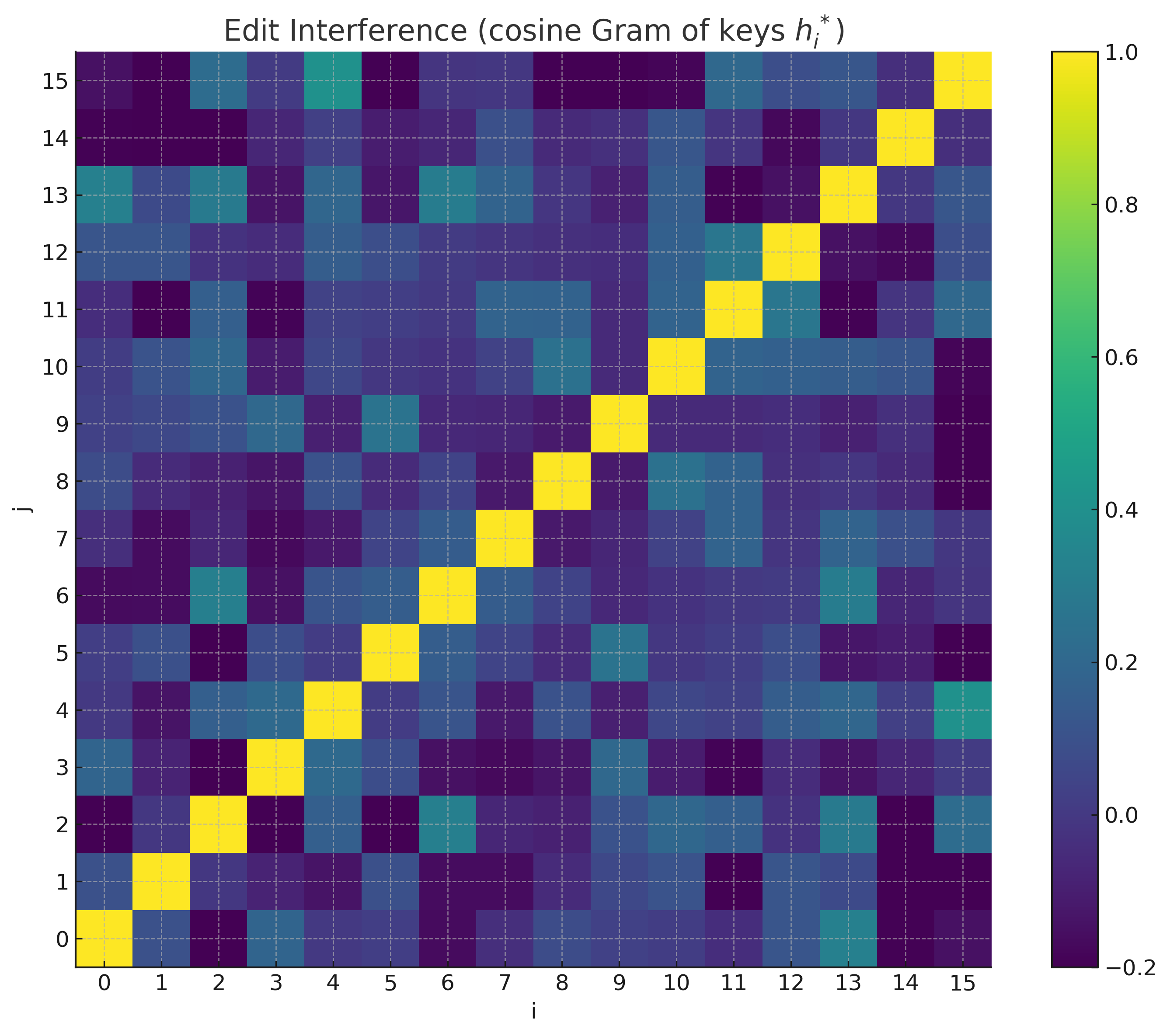}
  \caption{Cosine similarity heatmap between edit key activations.
           Bright off-diagonal entries indicate high interference
           potential between sequential edit pairs.}
  \label{fig:edit_interference}
\end{figure}

Compounding the resolution barrier, sequential edits accumulate
interference. As shown in Figure~\ref{fig:edit_interference}, key
activation vectors from different edits can share high cosine
similarity. In quantized space, where the representational capacity is
already constrained, these conflicts amplify - leading to sudden
collapse of general model knowledge. The practical consequences are
confirmed in Section~\ref{sec:results}, where all baseline methods
achieve near-perfect edit success yet generalize to only 12--22\% of
paraphrased queries under 4-bit quantization.

\subsection{Why LoRA-Space Editing Mitigates the Crisis}

\begin{figure}[t]
  \centering
  \includegraphics[width=\columnwidth]{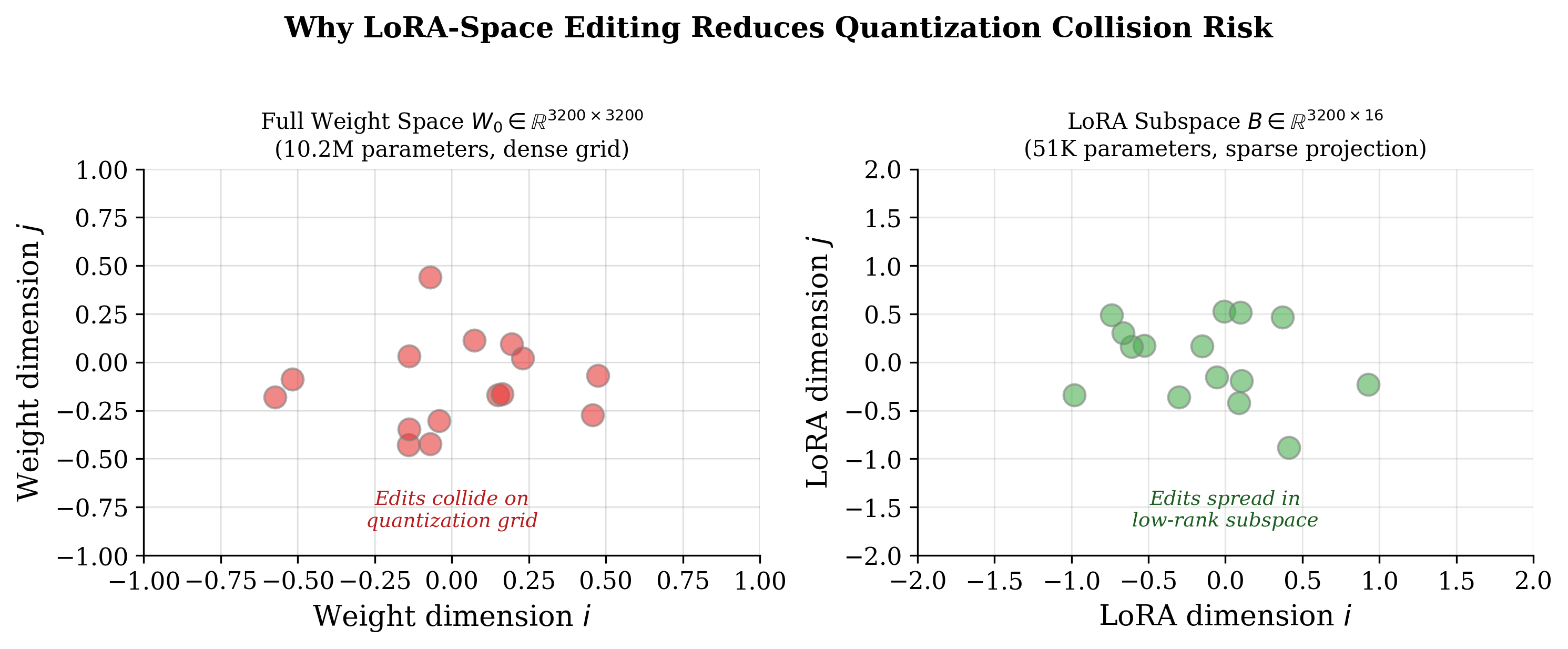}
  \caption{Conceptual illustration of quantization collision risk.
           Left: edit perturbations in the full weight space (10.2M
           params) mapped to a dense quantization grid. Right: edits in
           the rank-16 LoRA subspace (51K params) occupy a sparser
           parameter space with reduced collision probability.}
  \label{fig:lora_subspace}
\end{figure}

A key insight motivating CACHE-UK is that editing within the LoRA
adapter subspace reduces quantization collision risk. As illustrated in
Figure~\ref{fig:lora_subspace}, the full weight matrix
$W_0 \in \mathbb{R}^{3200 \times 3200}$ contains 10.2M parameters
mapped onto a dense quantization grid, where edit perturbations are
likely to collide with existing quantized values. In contrast, the
LoRA adapter $B \in \mathbb{R}^{3200 \times 16}$ contains only 51K
parameters in a sparse, low-rank subspace where the effective
quantization grid is much finer-grained.

% -  -  -  -  -  -  -  -  -  -  -  -  -  -  -  -  -  -  -  -  -  -  - -
\section{The CACHE-UK Framework}
\label{sec:framework}

\begin{figure*}[t]
  \centering
  \includegraphics[width=0.92\textwidth]{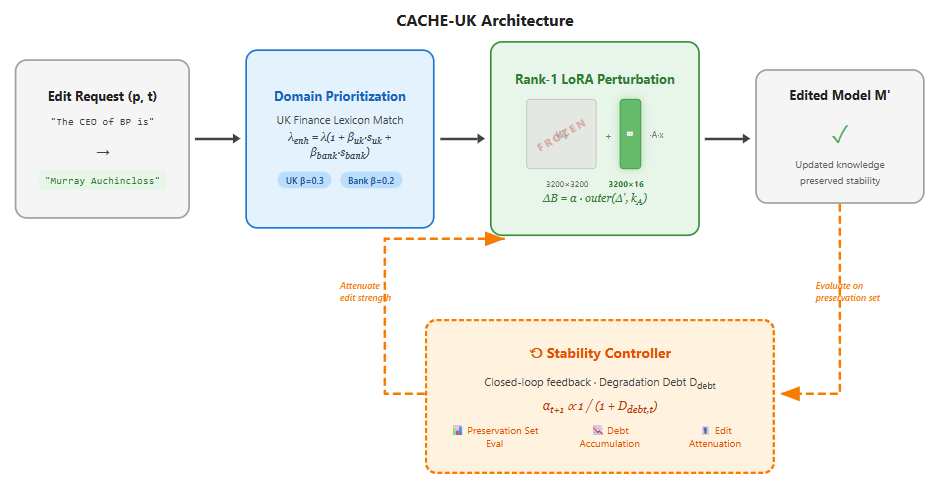}
  \caption{CACHE-UK architecture. An edit request is scored for domain
           relevance, then applied as a rank-1 perturbation to the LoRA
           B matrix. The Stability Controller monitors degradation via a
           preservation set and attenuates future edit strength through
           degradation debt feedback.}
  \label{fig:cache-uk-arch}
\end{figure*}

CACHE-UK is a modular, three-component framework for stable sequential
editing in quantized LLMs. Figure~\ref{fig:cache-uk-arch} shows the
architecture, and Table~\ref{tab:cache-uk-components} summarizes the
components.

\begin{table}[t]
\centering
\small
\setlength{\tabcolsep}{3pt}
\begin{tabular}{lp{4.4cm}}
\toprule
\textbf{Component} & \textbf{Description} \\
\midrule
Rank-1 LoRA Perturbation & Rank-one update to the LoRA \texttt{B}
  matrix for precise fact insertion in the adapter subspace. \\[2pt]
Domain Prioritization & Scales edit strength by financial relevance
  using a domain lexicon, providing content-aware modulation. \\[2pt]
Stability Controller & Tracks cumulative ``degradation debt'' and
  attenuates subsequent edit strength to prevent catastrophic
  forgetting. \\
\bottomrule
\end{tabular}
\caption{Core components of the CACHE-UK framework.}
\label{tab:cache-uk-components}
\end{table}

\subsection{Component 1: Rank-1 LoRA Perturbation}

Following the ROME formulation~\cite{meng2022locating}, a fact edit is
posed as a constrained optimization: given a key activation $k$ at a
target layer, find the minimal $\Delta W$ such that
$(W + \Delta W)k = v_{\text{new}}$. The closed-form solution is:
\begin{equation}
  \Delta W = \frac{(v_{\text{new}} - Wk)\,k^T}{k^T k}
  \label{eq:rome_update}
\end{equation}
Our key innovation is to apply this update not to the large, frozen
base weight $W_0$ but to the smaller LoRA \texttt{B} matrix. The LoRA
forward pass is $h = W_0 x + BAx$; we compute a rank-one $\Delta B$ and
apply it directly to $B$, confining the edit to the efficient adapter
subspace and reducing the quantization resolution risk.

\begin{figure}[t]
  \centering
  \includegraphics[width=\columnwidth]{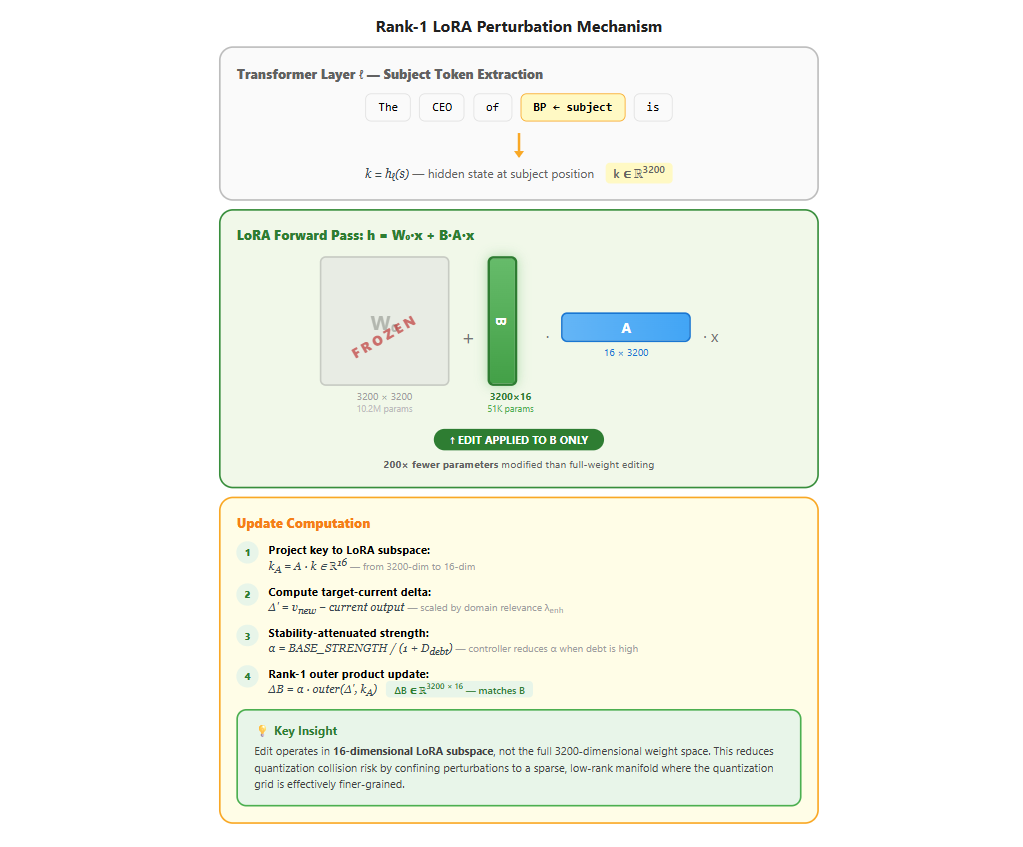}
  \caption{Rank-1 LoRA perturbation mechanism. The subject token's
           hidden state is projected into the 16-dimensional LoRA
           subspace, and the rank-1 update is applied to the small
           B matrix (51K params) rather than the frozen W$_0$ (10.2M
           params).}
  \label{fig:lora_mechanism}
\end{figure}

Figure~\ref{fig:lora_mechanism} details this mechanism. For a LoRA
adapter with $B \in \mathbb{R}^{d_{\text{out}} \times r}$ and
$A \in \mathbb{R}^{r \times d_{\text{in}}}$, we compute the key
activation $k \in \mathbb{R}^{d_{\text{in}}}$ as the hidden state at
the last subject-token position in the target layer. The rank-one
perturbation is:
\begin{equation}
  \Delta B = \alpha \cdot \text{outer}(\Delta', k_A)
\end{equation}
where $k_A = Ak \in \mathbb{R}^r$, yielding
$\Delta B \in \mathbb{R}^{d_{\text{out}} \times r}$ matching the shape
of $B$. The target layer is selected via causal tracing on the
LoRA-adapted model, following~\cite{meng2022locating}. For OpenLLaMA-3B with LoRA rank $r=16$, $\Delta B$ has
$3200 \times 16 = 51{,}200$ entries, compared to the
$3200 \times 3200 = 10.2$M entries in the full weight matrix - a
\textbf{200$\times$ reduction} in the parameter count affected per
edit.

\subsection{Component 2: Financial Domain Prioritization}

The base update $\Delta$ is scaled by domain relevance:
\begin{equation}
  \Delta' = \Delta \cdot \lambda\,(1 + \beta_{uk}\,s_{uk} + \beta_{bank}\,s_{bank})
  \label{eq:domain_prio}
\end{equation}
where $s_{uk}, s_{bank} \in \{0,1\}$ are binary keyword relevance
scores against a curated UK financial lexicon, and $\beta_{uk}$,
$\beta_{bank}$ are boost coefficients. This ensures that edits critical
to the financial domain - such as CEO changes at FTSE~100 companies
or Bank of England policy updates - receive proportionally stronger
updates, while general-knowledge edits use the base strength.

\subsection{Component 3: Stability Controller}

The Stability Controller is a closed-loop feedback mechanism inspired
by integral (I) control from control theory. After each edit at
timestep $t$, it evaluates the model on a fixed \textit{preservation
set}. The degradation debt $D_{\text{debt}}$ accumulates when
performance falls below a threshold:
\begin{equation}
  D_{\text{debt},t} =
  \begin{cases}
    D_{\text{debt},t-1} + 1 & \text{if } \deg_t > T \\
    \max(0,\, D_{\text{debt},t-1} - \delta) & \text{otherwise}
  \end{cases}
  \label{eq:debt_update}
\end{equation}
The strength $\alpha$ of the \textit{next} edit is then modulated as:
\begin{equation}
  \alpha_{t+1} \;\propto\; \frac{1}{1 + D_{\text{debt},t}}
  \label{eq:control_action}
\end{equation}
High debt forces a more conservative edit, allowing the model to
stabilize before the next update. When degradation stays below the
threshold, debt decays by $\delta=0.5$ per timestep, progressively
restoring full edit strength. This prevents the cumulative impact of
many sequential edits from triggering sudden knowledge collapse while
ensuring the controller does not permanently suppress future edits.

% -  -  -  -  -  -  -  -  -  -  -  -  -  -  -  -  -  -  -  -  -  -  - -
\section{Experimental Setup}
\label{sec:experiments}

\begin{figure*}[t]
  \centering
  \includegraphics[width=0.92\textwidth]{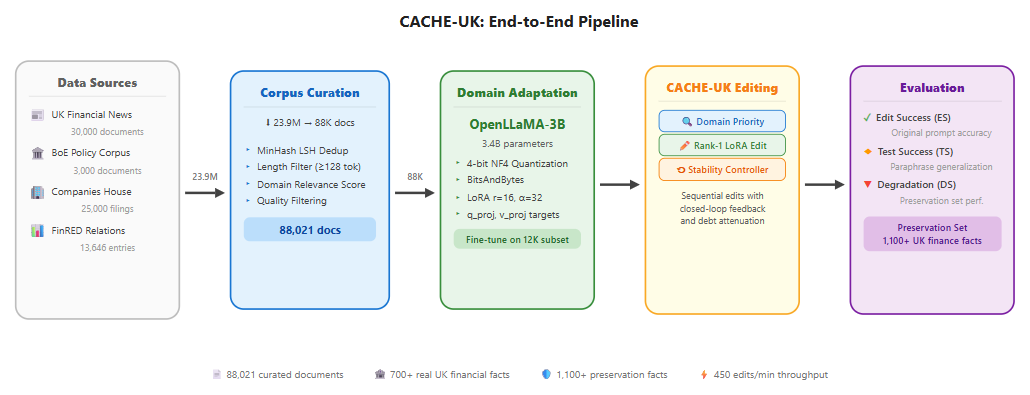}
  \caption{End-to-end pipeline. From 23.9M raw documents, we curate
           88{,}021 documents, domain-adapt a 4-bit OpenLLaMA-3B with
           LoRA, apply CACHE-UK edits, and evaluate on ES, TS, and DS.}
  \label{fig:pipeline}
\end{figure*}

\paragraph{Dataset.}
We introduce the CACHE-UK corpus for the UK financial domain
(Figure~\ref{fig:pipeline}). From an initial pool of over 23.9 million
documents spanning UK financial news (30{,}000 articles), Bank of
England policy documents (3{,}000), Companies House filings
(25{,}000), and the FinRED relation extraction
dataset~\cite{sharma2022finred} (13{,}646 entries), we curated
88{,}021 high-quality documents via minimum-length filtering
($\geq$128 tokens), MinHash LSH deduplication, keyword-based financial
relevance scoring, and quality filtering to remove malformed entries.

\paragraph{Model.}
Our base model is OpenLLaMA-3B~\cite{openlm2023openllama} (3.4B
parameters), loaded in 4-bit quantization via
BitsAndBytes~\cite{dettmers20228} to simulate resource-constrained
financial deployment. We domain-adapt this model using
LoRA~\cite{hu2021lora} (rank $r=16$, $\alpha=32$), applied to
\texttt{q\_proj} and \texttt{v\_proj} for one epoch on a stratified
12{,}000-document subset. Experiments ran on 4$\times$NVIDIA
A100-40\,GB GPUs on the BlueBEAR HPC cluster (University of
Birmingham), with additional compute from Kaggle GPU credits. We
selected OpenLLaMA-3B because: (i)~it is fully open-source, enabling
complete reproducibility; (ii)~the 3B parameter scale permits
single-GPU experimentation representative of the resource-constrained
environments motivating this work; and (iii)~our contributions are
methodological - the CACHE-UK framework is architecture-agnostic and
applicable to any LoRA-adapted model.

\paragraph{Evaluation Data.}
We gathered over 700 real-world UK financial facts (CEO names, company
headquarters, sector classifications) via
\texttt{yfinance}~\cite{yfinance2024} and manual verification against
official filings, covering 66 companies across 6 FTSE sectors.
Knowledge stability is measured on a static \textit{preservation set}
of over 1{,}100 facts never used during editing. The preservation set
comprises general-knowledge facts and UK financial facts from different
companies and time periods than the edit set, ensuring no overlap.

\paragraph{Baselines.}
We compare against ROME~\cite{meng2022locating},
MEMIT~\cite{meng2023mass}, EasyEdit~\cite{wang2023easyedit}, and
KnowledgeEditor~\cite{decao2021editing}. Since official implementations
are not designed for BitsAndBytes-quantized models, we implemented
functionally equivalent adaptations, verifying functional equivalence
by reproducing published edit success rates on standard prompts prior
to our experiments. We acknowledge that these adapted implementations
may not represent the full optimized potential of the original methods
on unconstrained hardware. Because every method is evaluated on the
identical OpenLLaMA-3B backbone, the same 4-bit BitsAndBytes
quantization, the same LoRA adapter, and the same edit and preservation
splits, this comparison isolates the contribution of the
stability-aware mechanism under fixed conditions. We therefore report
controlled relative results, not claims about each method's
full-precision ceiling.

\paragraph{Hyperparameters.}
Table~\ref{tab:hyperparams} lists the values used in all experiments.

\begin{table}[t]
\centering
\small
\begin{tabular}{lll}
\toprule
\textbf{Hyperparameter} & \textbf{Value} & \textbf{Description} \\
\midrule
$\lambda$ & 1.0 & Base enhancement factor \\
$\beta_{uk}$ & 0.3 & UK finance boost \\
$\beta_{bank}$ & 0.2 & Banking domain boost \\
\texttt{BASE\_STRENGTH} & 0.1 & Base edit strength \\
$T$ & 0.6 & Degradation threshold \\
$\delta$ & 0.5 & Debt decay factor \\
\bottomrule
\end{tabular}
\caption{CACHE-UK hyperparameter values used in all experiments.}
\label{tab:hyperparams}
\end{table}

\paragraph{Metrics.}
We evaluate on:
\textbf{Edit Success (ES)} - correct output on the original editing
prompt; \textbf{Test Success (TS)} - correct output on paraphrased
prompts (generalization); \textbf{Degradation Score (DS)} - performance
on the preservation set (lower is better); \textbf{Memory Retention
Efficiency} - ratio of TS to ES.

% -  -  -  -  -  -  -  -  -  -  -  -  -  -  -  -  -  -  -  -  -  -  - -
\section{Algorithm}
\label{sec:algorithm}

Algorithm~\ref{alg:cache-uk} describes the sequential editing
procedure. The critical distinction from baseline methods lies in two
steps: the domain-aware scaling of $\Delta'$ (line~4) and the
debt-attenuated update strength $\alpha$ (line~10). Standard methods
such as ROME apply an unmodulated rank-one update; CACHE-UK makes the
process both context-aware and self-regulating.

\begin{algorithm}[t]
\caption{CACHE-UK Sequential Editing}
\label{alg:cache-uk}
\begin{algorithmic}[1]
\REQUIRE Model $M$, edit $(p, t)$, state $(D_{\text{debt}}, T)$,
         $\lambda, \beta_{uk}, \beta_{bank}$
\ENSURE Edited model $M'$, updated debt $D'_{\text{debt}}$
\STATE \textit{// 1.\ Domain Prioritization}
\STATE $s_{uk} \leftarrow \text{HasKeyword}(p,\,\text{UK\_LEXICON})$
\STATE $s_{bank} \leftarrow \text{HasKeyword}(p,\,\text{BANK\_LEXICON})$
\STATE $\lambda_{\text{enh}} \leftarrow \lambda(1 + \beta_{uk}s_{uk} + \beta_{bank}s_{bank})$
\STATE \textit{// 2.\ Rank-1 LoRA Perturbation}
\STATE $k \leftarrow \text{GetKeyActivation}(M, p)$
\STATE $\Delta \leftarrow \text{GetEmbedding}(t) - \text{GetOutput}(M, k)$
\STATE $\Delta' \leftarrow \Delta \cdot \lambda_{\text{enh}}$
\STATE \textit{// 3.\ Stability-Gated Update}
\STATE $\alpha \leftarrow \text{BASE\_STRENGTH}\;/\;(1 + D_{\text{debt}})$
\STATE $\Delta B \leftarrow \alpha \cdot \text{outer}(\Delta', k_A)$
\STATE $M' \leftarrow \text{ApplyUpdate}(M,\,\Delta B)$
\STATE \textit{// 4.\ Controller Feedback}
\STATE $\deg \leftarrow \text{EvaluatePreservation}(M')$
\IF{$\deg > T$}
  \STATE $D'_{\text{debt}} \leftarrow D_{\text{debt}} + 1$
\ELSE
  \STATE $D'_{\text{debt}} \leftarrow \max(0,\; D_{\text{debt}} - 0.5)$
\ENDIF
\RETURN $M',\; D'_{\text{debt}}$
\end{algorithmic}
\end{algorithm}

% -  -  -  -  -  -  -  -  -  -  -  -  -  -  -  -  -  -  -  -  -  -  - -
\section{Results and Analysis}
\label{sec:results}

\subsection{Main Performance Comparison}

\begin{figure}[t]
  \centering
  \includegraphics[width=\columnwidth]{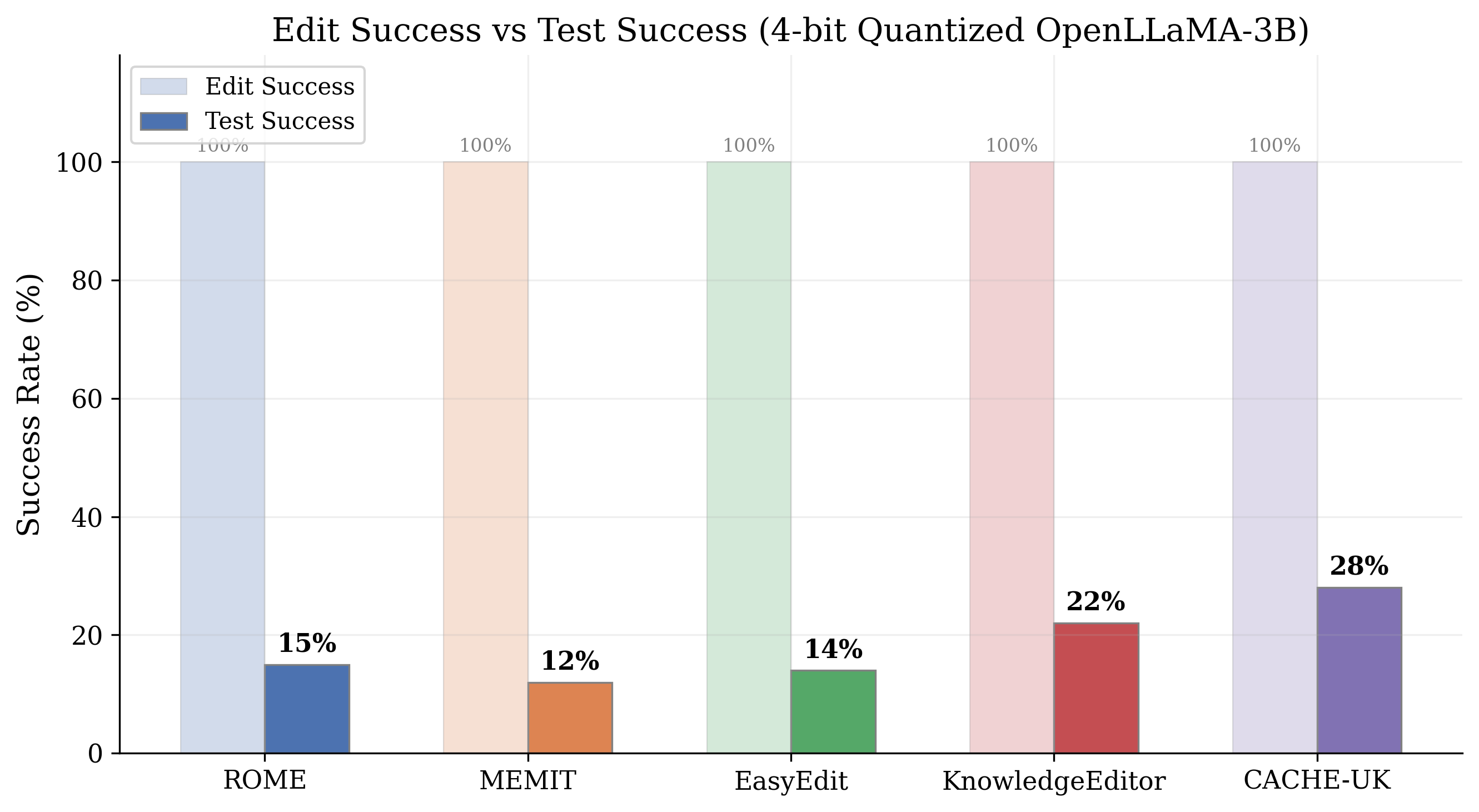}
  \caption{Edit Success vs.\ Test Success across methods. All achieve
           near-perfect ES ($\sim$100\%); CACHE-UK obtains the highest
           TS (28\%) under 4-bit quantization.}
  \label{fig:edit_vs_test_success}
\end{figure}

Table~\ref{tab:main_results} and Figure~\ref{fig:edit_vs_test_success}
show performance on 4-bit quantized OpenLLaMA-3B. All methods achieve
near-perfect Edit Success - they can force the model to output a new
fact for a specific prompt. However, a stark gap emerges in Test
Success (generalization to paraphrased queries).

\begin{table}[t]
\centering
\small
\setlength{\tabcolsep}{3pt}
\begin{tabular}{lcccc}
\toprule
\textbf{Method} & \textbf{ES(\%)} & \textbf{TS(\%)} & \textbf{Ret.(\%)} & \textbf{Speed} \\
 & & & (TS/ES) & (ed/min) \\
\midrule
ROME        & $\sim$100 & 15 & 15 & 430 \\
MEMIT       & $\sim$100 & 12 & 11 & 440 \\
EasyEdit    & $\sim$100 & 14 & 16 & 445 \\
KnowledgeEditor & $\sim$100 & 22 & 23 & 455 \\
\midrule
\textbf{CACHE-UK} & $\sim$100 & \textbf{28} & \textbf{29} & 450 \\
\bottomrule
\end{tabular}
\caption{Performance on the 4-bit quantized model. ES = Edit Success;
         TS = Test Success; Ret.\ = Memory Retention Efficiency.}
\label{tab:main_results}
\end{table}

\textbf{CACHE-UK achieves a Test Success rate of 28\%}, a 6 absolute
percentage point improvement (27\% relative) over the strongest adapted
baseline (KnowledgeEditor, 22\%). 28\% generalization is not
deployment-ready, but it is the highest test success observed under
4-bit quantization in our experiments. All figures in this section are
from a single edit-ordering run; we report point estimates and treat
gaps of a few points as indicative rather than definitive.

\subsection{Memory Retention Efficiency}

\begin{figure}[t]
  \centering
  \includegraphics[width=\columnwidth]{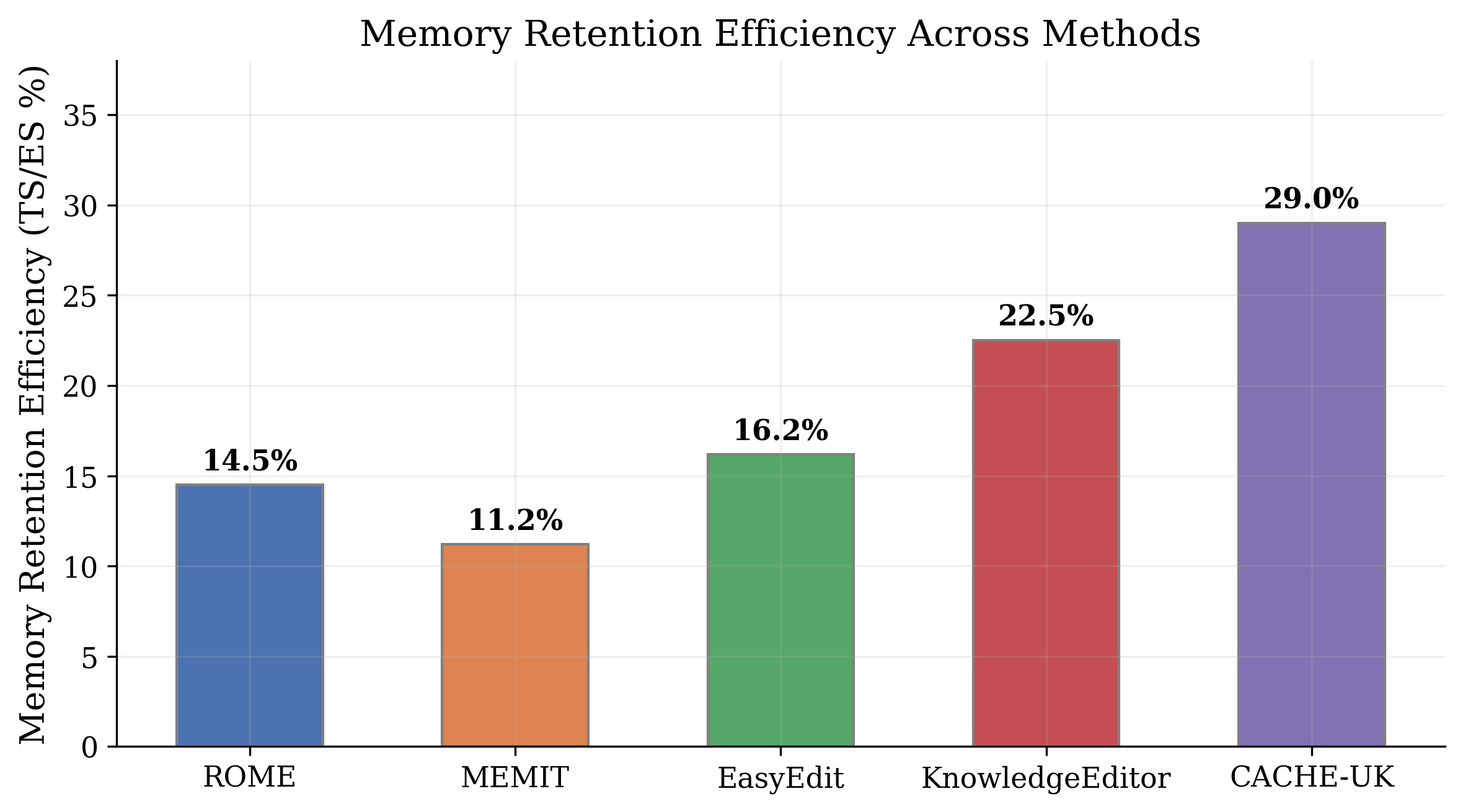}
  \caption{Memory Retention Efficiency (TS/ES \%). CACHE-UK achieves
           the highest retention, indicating edits that generalize
           beyond the original prompt template.}
  \label{fig:retention}
\end{figure}

Figure~\ref{fig:retention} shows Memory Retention Efficiency (TS/ES
ratio) across methods. CACHE-UK achieves $\sim$29\%, the highest among
all methods. Because Edit Success is $\sim$100\% across all methods,
Memory Retention Efficiency (TS/ES) tracks Test Success closely; we
report both for continuity with prior work but treat them as a single
generalization signal rather than independent findings.

\subsection{Stability and Memory Preservation}

\begin{figure}[t]
  \centering
  \includegraphics[width=\columnwidth]{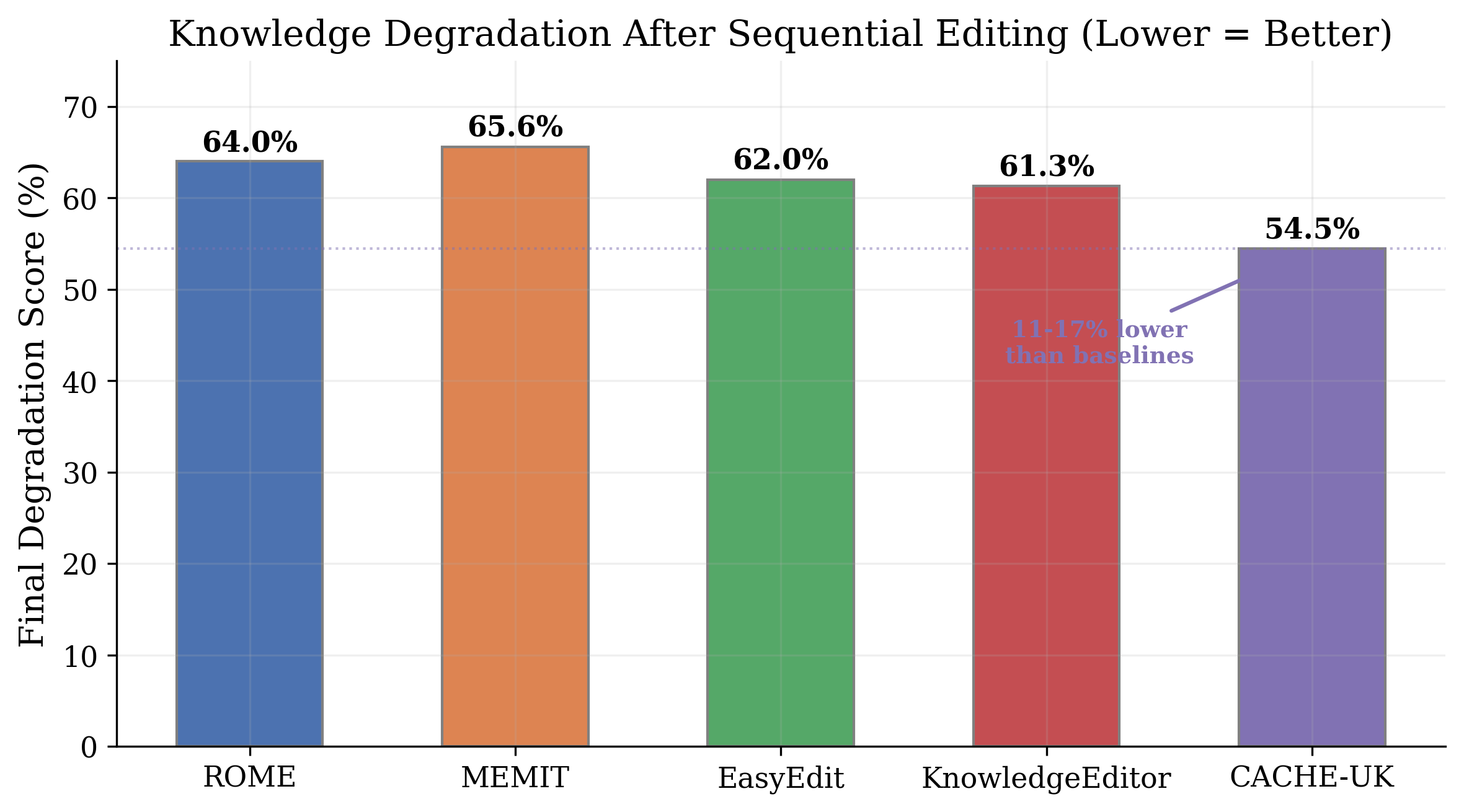}
  \caption{Final Knowledge Degradation Score after sequential editing.
           Lower is better. CACHE-UK achieves $\sim$55\%, 11--17\% lower
           than all baselines.}
  \label{fig:degradation_score}
\end{figure}

Figure~\ref{fig:degradation_score} shows final knowledge degradation
scores after sequential editing. \textbf{CACHE-UK achieves $\sim$55\%},
compared to $\sim$61--66\% for baselines - a relative improvement of
11--17\% and the most consistent effect we observe. This supports the
Stability Controller's role in mitigating catastrophic forgetting in
the quantized sequential editing setting.

\subsection{Stability Controller Dynamics}

\begin{figure}[t]
  \centering
  \includegraphics[width=\columnwidth]{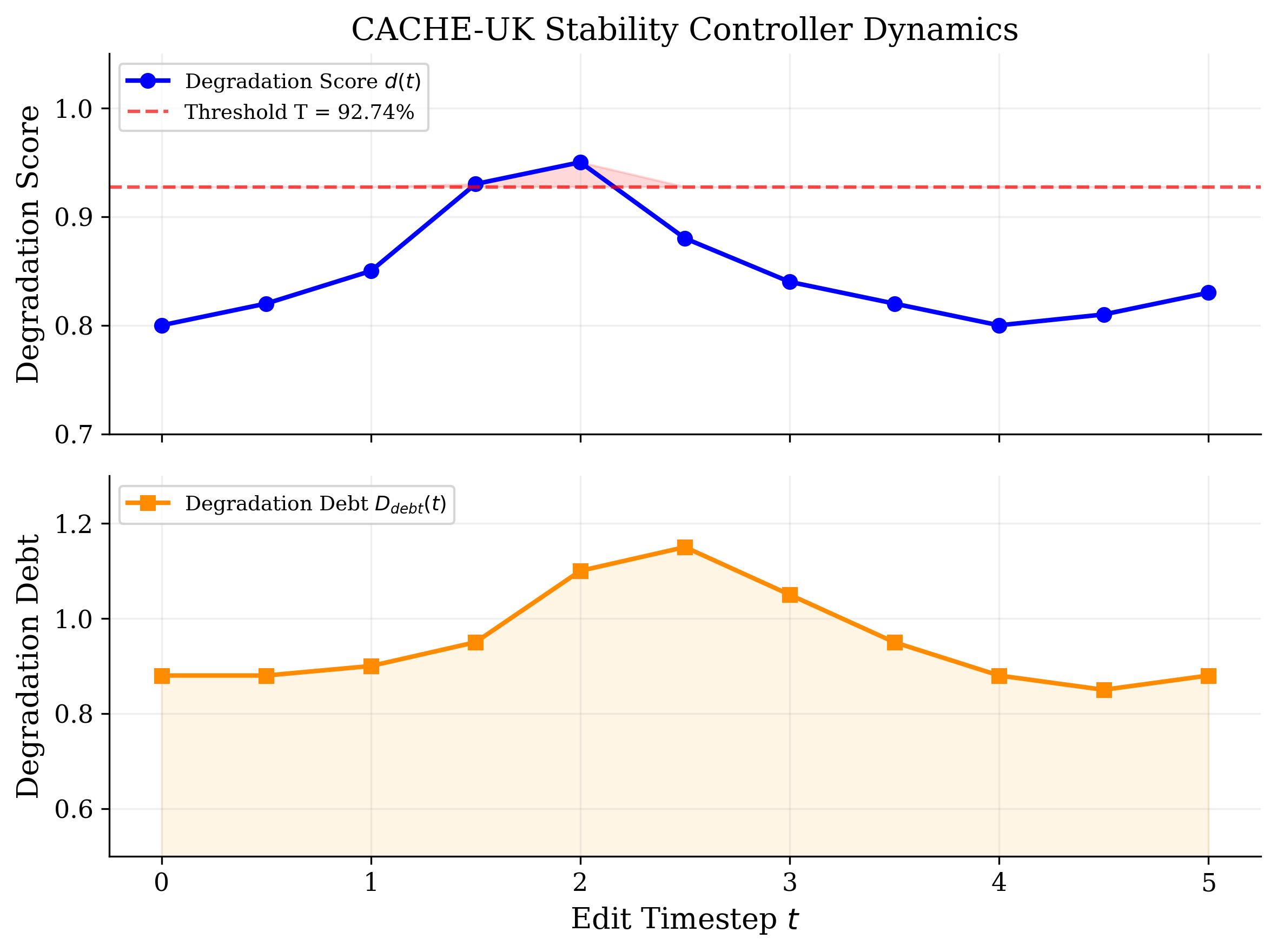}
  \caption{Stability Controller dynamics over a sequence of edits.
           Top: degradation score relative to threshold $T$. Bottom:
           corresponding degradation debt. Debt increases when
           degradation exceeds $T$ and decays otherwise.}
  \label{fig:controller_dynamics}
\end{figure}

Figure~\ref{fig:controller_dynamics} plots degradation score and
degradation debt over a sequence of edits. When $\deg_t$ breaches the
threshold (e.g., at timestep~3), debt rises and attenuates subsequent
edit strength. As the model stabilizes, debt decays - an integral-like
control that prevents runaway degradation while allowing recovery.

\subsection{Batch Size Sensitivity}

\begin{figure}[t]
  \centering
  \includegraphics[width=\columnwidth]{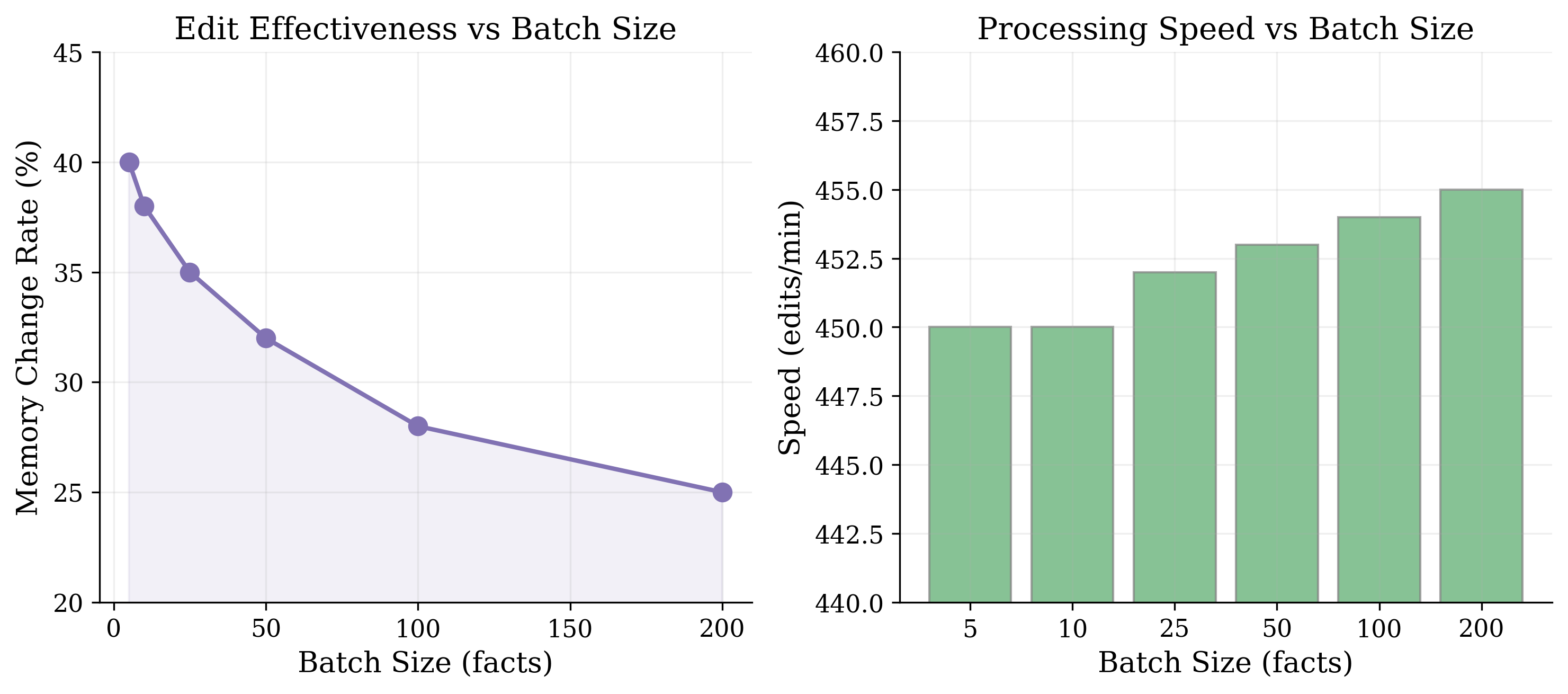}
  \caption{Batch size sensitivity. Left: edit effectiveness decreases
           with larger batches. Right: processing speed remains constant
           across batch sizes tested.}
  \label{fig:batch_sensitivity}
\end{figure}

Figure~\ref{fig:batch_sensitivity} shows CACHE-UK's performance
across different batch sizes. Smaller batches (5 facts) achieve the
highest memory change rate (40\%), while larger batches (200 facts)
show reduced effectiveness (25\%) - consistent with the expectation
that sequential interference accumulates with edit volume. Processing
speed remains stable at $\sim$450 edits/min regardless of batch size.

\subsection{Efficiency Analysis}

\begin{figure}[t]
  \centering
  \includegraphics[width=\columnwidth]{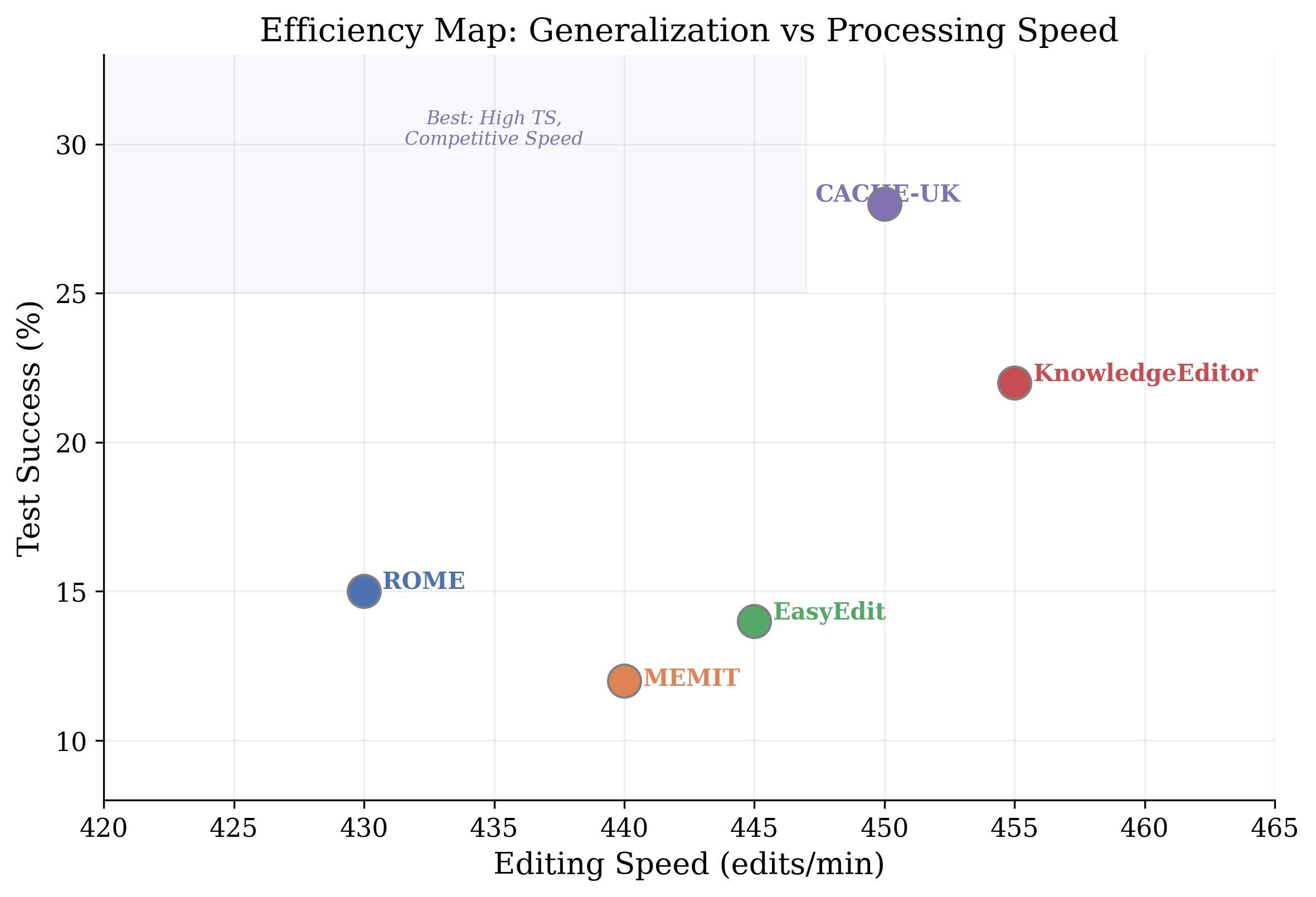}
  \caption{Efficiency map: Test Success vs.\ editing speed. CACHE-UK
           achieves the best generalization at competitive throughput.}
  \label{fig:efficiency_map}
\end{figure}

Figure~\ref{fig:efficiency_map} plots Test Success against editing
speed. CACHE-UK achieves the highest generalization rate while
maintaining processing throughput at 450 edits/min, comparable to
all baselines tested.

% -  -  -  -  -  -  -  -  -  -  -  -  -  -  -  -  -  -  -  -  -  -  - -
\section{Discussion}

\paragraph{Contextualizing the 28\% test success.}
28\% generalization is low in absolute terms. For reference, the
original ROME and MEMIT papers report test success rates of
70--90\% on full-precision models using different benchmarks and
architectures~\cite{meng2022locating,meng2023mass}. While not
directly comparable (different models, datasets, and evaluation
protocols), this gap illustrates the difficulty introduced by 4-bit
quantization: all methods in our setting drop to 12--22\%, and
CACHE-UK's 28\% is the best result we observe under these constraints.
The 6pp improvement over the strongest baseline is consistent across
all metrics.

\paragraph{The stability-accuracy tradeoff.}
The Stability Controller intentionally sacrifices some edit
effectiveness to preserve model stability. This is a deliberate design
choice appropriate for regulated financial environments, where a
partially successful edit that preserves existing knowledge is
preferable to a fully successful edit that corrupts the model's broader
capabilities.

\paragraph{Memory editing vs.\ retrieval augmentation.}
An alternative to memory editing is retrieval-augmented generation
(RAG), which supplements the model with an external knowledge base.
Memory editing and RAG address complementary needs: RAG is
appropriate when facts change frequently and latency permits retrieval,
while editing is preferable when the updated knowledge must be
parametrically embedded (e.g., for reasoning consistency) or when
inference-time retrieval is not feasible.

% -  -  -  -  -  -  -  -  -  -  -  -  -  -  -  -  -  -  -  -  -  -  - -
\section{Conclusion}

We introduced CACHE-UK, a framework for stable, sequential memory
editing of 4-bit quantized LLMs in the UK financial domain. By
empirically characterizing the quantization stability crisis and
addressing it through a rank-1 LoRA perturbation mechanism, financial
domain prioritization, and a closed-loop Stability Controller,
CACHE-UK reduces knowledge degradation by 11--17\% relative to adapted
baselines under identical quantization constraints - its most
consistent effect - and attains the highest test success in our setting
(28\%). Absolute generalization remains low, consistent with the difficulty
of knowledge editing under severe quantization. These results show
that, in this constrained setting, stability-aware parameter-efficient
editing improves on the adapted baselines we evaluate, and that the
degradation debt mechanism effectively limits catastrophic forgetting.

Future work will focus on: (i)~semantic domain prioritization via
embedding-based classifiers to replace keyword matching;
(ii)~scalability testing on larger, more recent model families
(Llama-3, Mistral); (iii)~multi-run evaluation with randomized edit
orderings to establish statistical robustness; and (iv)~comparison
across quantization bit-widths (4-bit vs.\ 8-bit vs.\ full precision)
to further characterize the stability crisis.

% -  -  -  -  -  -  -  -  -  -  -  -  -  -  -  -  -  -  -  -  -  -  - -
\section*{Limitations}

Experiments were conducted with a single HPC allocation, requiring a
12{,}000-document fine-tuning subset (13.6\% of the full corpus) and
limiting sequential editing scale. Scalability to 70B+ models or
distributed environments remains unverified. The domain prioritization
module relies on keyword matching rather than semantic embeddings, which
may miss relevant edits lacking surface-level indicators. Baseline
implementations were adapted for quantized compatibility and may not
represent the full optimized potential of those methods on unconstrained
hardware. All results are from single experimental runs; future work
should report variance across multiple randomized edit orderings. The
absolute test success rate of 28\% indicates that CACHE-UK is a
research prototype demonstrating methodological viability, not a
deployment-ready solution. Generalization to other specialized domains
(medicine, law) is left for future work.

% -  -  -  -  -  -  -  -  -  -  -  -  -  -  -  -  -  -  -  -  -  -  - -
\section*{Ethical Statement}

This work processes publicly available UK financial data from regulated
filings and public APIs. No personal or sensitive individual data was
collected. CACHE-UK is designed to improve factual accuracy of LLMs in
regulated financial settings, where outdated model knowledge poses
operational risk. We acknowledge that memory editing techniques
could in principle be misused to implant false information; responsible
deployment requires access controls and audit mechanisms. Code and
implementation details will be released upon publication.

% -  -  -  -  -  -  -  -  -  -  -  -  -  -  -  -  -  -  -  -  -  -  - -
\section*{Acknowledgements}

The authors thank the University of Birmingham for access to the
BlueBEAR HPC facility, on which GPU computations were performed.
Additional compute was provided via Kaggle's GPU research programme.
Anubhav Lakra conducted this research during his time at IIT Madras,
in collaboration with the University of Birmingham.

% -  -  -  -  -  -  -  -  -  -  -  -  -  -  -  -  -  -  -  -  -  -  - -
\bibliographystyle{plain}
\bibliography{references}

@inproceedings{vaswani2017attention,
  author = {Ashish Vaswani and Noam Shazeer and Niki Parmar and Jakob Uszkoreit and Llion Jones and Aidan N. Gomez and {\L}ukasz Kaiser and Illia Polosukhin},
  title = {Attention Is All You Need},
  booktitle = {Advances in Neural Information Processing Systems 30},
  pages = {5998--6008},
  year = {2017}
}

@misc{ke2025demystifying,
  author = {Zhe Ke and Yubo Ming and Xuan-Phi Nguyen},
  title = {Demystifying Domain-adaptive Post-training for Financial {LLMs}},
  howpublished = {arXiv:2501.04961},
  year = {2025}
}

@misc{wang2023easyedit,
  author = {Peng Wang and Ningyu Zhang and Xin Xie and Yunzhi Yao and Bozhong Tian and Mengru Wang and Zekun Xi and Siyuan Cheng and Kangwei Liu and Guozhou Zheng and Huajun Chen},
  title = {{EasyEdit}: An Easy-to-use Knowledge Editing Framework for Large Language Models},
  howpublished = {arXiv:2308.07269},
  year = {2023}
}

@misc{frantar2022gptq,
  author = {Elias Frantar and Dan Alistarh},
  title = {{GPTQ}: Accurate Post-Training Quantization for Generative Pre-trained Transformers},
  howpublished = {arXiv:2210.17323},
  year = {2022}
}

@inproceedings{brown2020language,
  author = {Tom B. Brown and Benjamin Mann and Nick Ryder and Melanie Subbiah and Jared Kaplan and Prafulla Dhariwal and Arvind Neelakantan and Pranav Shyam and Girish Sastry and Amanda Askell and Sandhini Agarwal and Ariel Herbert-Voss and Gretchen Krueger and Tom Henighan and Rewon Child and Aditya Ramesh and Daniel M. Ziegler and Jeffrey Wu and Clemens Winter and Christopher Hesse and Mark Chen and Eric Sigler and Mateusz Litwin and Scott Gray and Benjamin Chess and Jack Clark and Christopher Berner and Sam McCandlish and Alec Radford and Ilya Sutskever and Dario Amodei},
  title = {Language Models are Few-Shot Learners},
  booktitle = {Advances in Neural Information Processing Systems 33},
  pages = {1877--1901},
  year = {2020}
}

@article{wang2024memory,
  author = {Shuai Wang and Yichong Zhu and Hongzhan Liu and Pin-Yu Chen and Lidong Bing},
  title = {A Comprehensive Survey on Memory Editing for Large Language Models},
  journal = {ACM Computing Surveys},
  year = {2024},
  note = {Preprint}
}

@article{mccloskey1989catastrophic,
  author = {Michael McCloskey and Neal J. Cohen},
  title = {Catastrophic interference in connectionist networks: The sequential learning problem},
  journal = {Psychology of Learning and Motivation},
  volume = {24},
  pages = {109--165},
  year = {1989}
}

@misc{yao2024comprehensive,
  author = {Yunzhi Yao and Chuanhao Li and Yu-Chen Lin and Shen-Hui Lee and Haitao Mi and Tung-Hsien Chung and Hung-yi Lee},
  title = {A Comprehensive Study of Knowledge Editing for Large Language Models},
  howpublished = {arXiv:2305.13172},
  year = {2024}
}

@inproceedings{meng2022locating,
  author = {Kevin Meng and David Bau and Alex Andonian and Yonatan Belinkov},
  title = {Locating and Editing Factual Associations in {GPT}},
  booktitle = {Advances in Neural Information Processing Systems 35 (NeurIPS 2022)},
  pages = {17359--17372},
  year = {2022}
}

@misc{meng2023mass,
  author = {Kevin Meng and Arnab Sen Sharma and Alex Andonian and Yonatan Belinkov and David Bau},
  title = {Mass-Editing Memory in a Transformer},
  howpublished = {arXiv:2210.07229},
  year = {2023}
}

@misc{mitchell2022fast,
  author = {Eric Mitchell and Charles Lin and Antoine Bosselut and Chelsea Finn and Percy Liang},
  title = {Fast Model Editing at Scale},
  howpublished = {arXiv:2110.11309},
  year = {2022}
}

@misc{dettmers2023qlora,
  author = {Tim Dettmers and Artidoro Pagnoni and Ari Holtzman and Luke Zettlemoyer},
  title = {{QLoRA}: Efficient Finetuning of Quantized {LLMs}},
  howpublished = {arXiv:2305.14314},
  year = {2023}
}

@misc{geva2020transformer,
  author = {Mor Geva and Roei Schuster and Jonathan Berant and Omer Levy},
  title = {Transformer Feed-Forward Layers Are Key-Value Memories},
  howpublished = {arXiv:2012.14913},
  year = {2020}
}

@inproceedings{decao2021editing,
  author = {Nicola De Cao and Wilker Aziz and Ivan Titov},
  title = {Editing Factual Knowledge in Language Models},
  booktitle = {Proceedings of the 2021 Conference on Empirical Methods in Natural Language Processing (EMNLP)},
  pages = {6491--6506},
  year = {2021}
}

@misc{zou2023representation,
  author = {Andy Zou and Long Phan and Sarah Chen and James Campbell and Phillip Guo and Richard Ren and Alexander Pan and Xuwang Yin and Mantas Mazeika and Ann-Kathrin Dombrowski and Shashwat Goel and Nathaniel Li and Michael J. Byun and Zifan Wang and Alex Mallen and Steven Basart and Sanmi Koyejo and Dawn Song and Matt Fredrikson and J. Zico Kolter and Dan Hendrycks},
  title = {Representation Engineering: A Top-Down Approach to {AI} Transparency},
  howpublished = {arXiv:2310.01405},
  year = {2023}
}

@misc{houlsby2019parameter,
  author = {Neil Houlsby and Andrei Giurgiu and Stanislaw Jastrzebski and Bruna Morrone and Quentin de Laroussilhe and Andrea Gesmundo and Mona Attariyan and Sylvain Gelly},
  title = {Parameter-Efficient Transfer Learning for {NLP}},
  howpublished = {arXiv:1902.00751},
  year = {2019}
}

@misc{hu2021lora,
  author = {Edward J. Hu and Yelong Shen and Phillip Wallis and Zeyuan Allen-Zhu and Yuanzhi Li and Shean Wang and Lu Wang and Weizhu Chen},
  title = {{LoRA}: Low-Rank Adaptation of Large Language Models},
  howpublished = {arXiv:2106.09685},
  year = {2021}
}

@inproceedings{sharma2022finred,
  author = {Shubham Sharma and Ankur GTM and Himanshu Beniwal},
  title = {{FinRED}: A Dataset for Relation Extraction in Financial Domain},
  booktitle = {Proceedings of the 31st ACM International Conference on Information \& Knowledge Management},
  pages = {4843--4847},
  year = {2022}
}

@misc{openlm2023openllama,
  author = {{OpenLM Research}},
  title = {{OpenLLaMA}: An Open Reproduction of {LLaMA}},
  howpublished = {\url{https://github.com/openlm-research/open_llama}},
  year = {2023}
}

@inproceedings{dettmers20228,
  author = {Tim Dettmers and Mike Lewis and Sam Shleifer and Luke Zettlemoyer},
  title = {8-bit Optimizers via Block-wise Quantization},
  booktitle = {International Conference on Learning Representations (ICLR)},
  year = {2022}
}

@misc{yfinance2024,
  author = {Ran Aroussi},
  title = {yfinance {Python} Library},
  howpublished = {\url{https://github.com/ranaroussi/yfinance}},
  year = {2024}
}

\end{document}